\documentclass{article}

\usepackage{microtype}
\usepackage{graphicx}
\usepackage{subfigure}
\usepackage{booktabs} % for professional tables

\usepackage{hyperref}

\usepackage[accepted]{icml2025}

\usepackage{amsmath}
\usepackage{amssymb}
\usepackage{mathtools}
\usepackage{amsthm}
\usepackage{bbm}
\usepackage{bm}

\usepackage[capitalize,noabbrev]{cleveref}

\theoremstyle{plain}

\theoremstyle{definition}

\theoremstyle{remark}

\usepackage[textsize=tiny]{todonotes}

\icmltitlerunning{Causal Foundation Models}

\begin{document}

\twocolumn[
\icmltitle{Causal Foundation Models: \\ Disentangling Physics from Instrument Properties}



\icmlsetsymbol{equal}{*}

\begin{icmlauthorlist}
\icmlauthor{Jeroen Audenaert}{equal,mit}
\icmlauthor{Daniel Muthukrishna}{equal,mit}
\icmlauthor{Paul F. Gregory}{mit}
\icmlauthor{David W. Hogg}{nyu,flat,max}
\icmlauthor{V. Ashley Villar}{harvard,iaifi}
\end{icmlauthorlist}

\icmlaffiliation{mit}{Massachusetts Institute of Technology, Camridge, MA, USA}
\icmlaffiliation{harvard}{Harvard University, Cambridge, MA, USA}
\icmlaffiliation{nyu}{New York University, New York, NY, USA}
\icmlaffiliation{flat}{Flatiron Institute, New York, NY, USA}
\icmlaffiliation{max}{Max-Planck-Institut f\"ur Astronomie, Heidelberg, Germany}
\icmlaffiliation{iaifi}{The NSF AI Institute for Artificial Intelligence and Fundamental Interactions}

\icmlcorrespondingauthor{Jeroen Audenaert}{jeroena@mit.edu}
\icmlcorrespondingauthor{Daniel Muthukrishna}{danmuth@mit.edu}

\icmlkeywords{Machine Learning, ICML}

\vskip 0.3in
]



\printAffiliationsAndNotice{\icmlEqualContribution} % otherwise use the standard text.

\begin{abstract}
Foundation models for structured time series data must contend with a fundamental challenge: observations often conflate the true underlying physical phenomena with systematic distortions introduced by measurement instruments. This entanglement limits model generalization, especially in heterogeneous or multi-instrument settings. We present a causally-motivated foundation model that explicitly disentangles physical and instrumental factors using a dual-encoder architecture trained with structured contrastive learning. Leveraging naturally occurring observational triplets (i.e., where the same target is measured under varying conditions, and distinct targets are measured under shared conditions) our model learns separate latent representations for the underlying physical signal and instrument effects. Evaluated on simulated astronomical time series designed to resemble the complexity of variable stars observed by missions like NASA's Transiting Exoplanet Survey Satellite (TESS), our method significantly outperforms traditional single-latent space foundation models on downstream prediction tasks, particularly in low-data regimes. These results demonstrate that our model supports key capabilities of foundation models, including few-shot generalization and efficient adaptation, and highlight the importance of encoding causal structure into representation learning for structured data.

\end{abstract}

\section{Introduction}
\label{sect:intro}

Observational datasets across scientific and industrial domains often conflate two distinct sources of variation: (i) the true underlying signal of interest, and (ii) distortions introduced by measurement tools, such as sensor drift, calibration offsets and environmental or observing conditions. This entanglement poses a fundamental challenge to building foundation models that can generalize across instruments, domains, or modalities.

In astronomy, the availability of petabyte-scale, open-access data \citep[e.g.,][]{mmu2024} has recently spurred a rapid development of foundation models \citep[e.g.,][]{Parker2024,Rizhko2024,Zhang2024,EuclidSiudek2025}. Time series foundation models are also emerging across broader domains \citep[e.g.,][]{goswami2024moment,das2024a}. However,  astrophysical signals are often deeply entangled with systematic instrumental effects. This entanglement limits the interpretability and generalization of learned representations. For example, in \citet{EuclidSiudek2025} the instrumental properties were found to be separated in the latent space. Fortunately, many astronomical surveys provide natural experimental structure: the same star is frequently observed under different instrument configurations, and the same configuration observes many stars. This recurring observational pattern offers a unique opportunity to disentangle underlying physical dynamics from instrumental signatures.

In this work, we leverage these structural properties to develop a foundation model that explicitly separates physical and instrumental factors. Our method is inspired by causal representation learning \citep{Scholkopf2016,Scholkopf2021} and contrastive learning \citep{Chen2020}. We validate our approach on a simulated dataset designed to resemble the complexity of variable star light curves (brightness over time) observed by the Transiting Exoplanet Survey Satellite \citep[TESS;][]{ricker2015}. By learning general-purpose, causally disentangled representations that support few-shot learning and transfer across conditions, our model exhibits key capabilities of foundation models for structured time series. We can train downstream tasks from either of our physics or instrument latent representations. The remainder of this paper is organized as follows: Section~\ref{sect:data} introduces the simulation framework. Section~\ref{sect:model} describes the model architecture and contrastive training strategy. Section~\ref{sect:results} evaluates the learned latent spaces and their utility in downstream prediction tasks.

\section{Data}
\label{sect:data}

We construct a simulated dataset of time series observations designed to emulate the causal structure of real-world astronomical surveys, where each observed light curve reflects both intrinsic stellar variability and systematic effects introduced by measurement instruments. This controlled setting enables direct evaluation of a model’s ability to disentangle physical and instrumental factors in the latent space.

Each synthetic observation $F_{\text{observed}}^{(n)}(t)$ is generated by modulating a true stellar signal $F_{\text{true}}^{(s)}(t)$ with an instrument-specific scale $S_m(t)$ and offset $O_m(t)$, followed by the addition of Gaussian noise:
\begin{equation}
F_{\text{observed}}^{(n)}(t) = \text{clip}_{[-1, 1]}\left(S_m(t) \cdot F_{\text{true}}^{(s)}(t) + O_m(t)\right) + \varepsilon,
\end{equation}

where $s$ and $m$ index the star and instrument, respectively, $\varepsilon \sim \mathcal{N}(0, 0.03^2)$ adds observational noise, and $\text{clip}_{[-1, 1]}$  bounds the signal amplitude to the range $[-1, 1]$.

The true stellar signal is modeled as a complex Fourier series:
\begin{equation}
F_{\text{true}}^{(s)}(t) = \text{Re} \left[ \sum_{k=1}^{K-1} \frac{\theta_{s,k} \cdot e^{\mathrm{i} \phi_{n,k-1}}}{k^{\alpha}} \cdot e^{\mathrm{i} 2\pi k t / L_s} \right],
\end{equation}
where $\theta_{s,k}$ are stellar parameters, $\phi_{n,k}$ are observation-specific phases, $\alpha$ controls the power-law decay of frequency components, and $L_s = T \cdot e^{\theta_{s,0}} \cdot \lambda$ is the star’s period modulated by a reduction factor $\lambda$.

Instrumental distortions are also constructed as Fourier series:
\begin{align}
S_m(t) = 1 + 0.05 \cdot \text{Re}\left[\sum_{j=0}^{M-1} \beta_{m,j} \cdot e^{\mathrm{i} \pi j t / T} \right], \\
O_m(t) = 0.05 \cdot \text{Re}\left[\sum_{j=0}^{M-1} \gamma_{m,j} \cdot e^{\mathrm{i} \pi j t / T} \right],
\end{align}

with $\beta_{m,j}, \gamma_{m,j} \sim \mathcal{N}(0,1)$ serving as instrument-specific parameters.

Each observation $n$ is uniquely defined by a star--instrument pair $(s, m)$. We generate large datasets with multiple repeated measurements of the same star under different instruments, and different stars observed with the same instrument. This structure mirrors the observing strategy of surveys like TESS, and is crucial to enabling the contrastive learning framework used in our method.

The resulting dataset contains 40,000 light curves, 2000 unique stars, 17 instrument configurations ($M=17$), and 13 stellar parameters ($K=13$). Each observation consists of $T=100$ time steps. This scale provides sufficient statistical diversity while allowing for efficient training and evaluation. Example simulations are illustrated in Fig.~\ref{fig:example-lc}.

% We construct training triplets consisting of (1) an anchor observation, (2) another observation of the same star with a different instrument, and (3) a different star observed with the same instrument. By applying contrastive losses in each latent space, we learn disentangled representations that are invariant to nuisance variation in the other factor.

\begin{figure}[!th]
\vskip 0.2in
\begin{center}
\centerline{\includegraphics[width=\columnwidth]{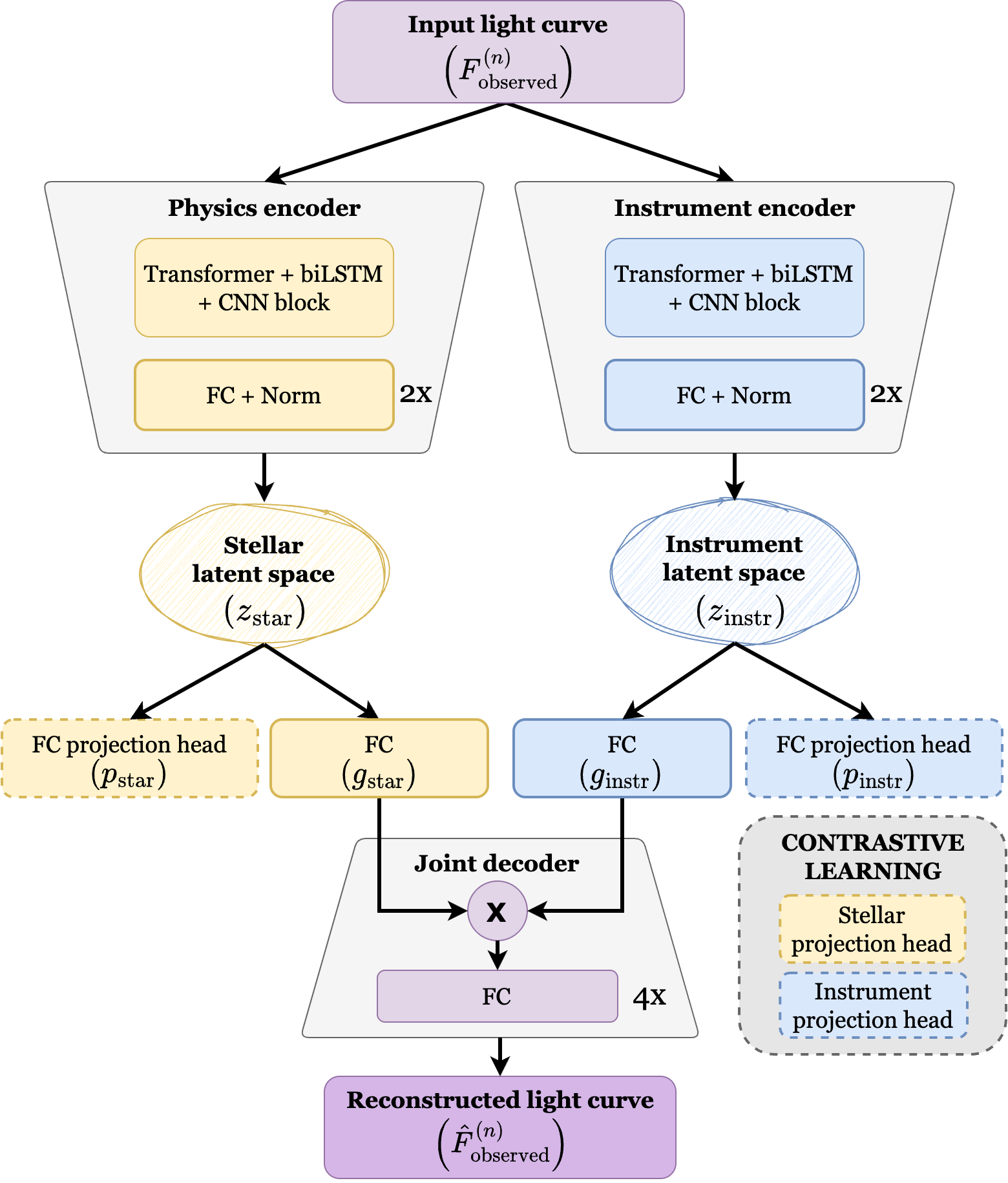}}
%\vskip -0.1in
\caption{Causal autoencoder architecture for light curve disentanglement. The model employs dual encoders (stellar and instrumental) that process the same input independently to learn separate latent representations. The decoder combines these representations through an element-wise multiplicative interaction. Projection heads (dashed lines) enable contrastive learning during training but are not used for inference.}
\label{fig:model-architecture}
\end{center}
\vskip -0.2in
\end{figure}

\section{Methods}
\label{sect:model}

We propose a causal foundation model for structured time series that disentangles physical properties of the observed system from systematic effects introduced by the measurement process. Our method builds on the assumption common in astronomical and other sensor-driven domains that observations are conditionally independent mixtures of underlying generative sources (e.g., stellar parameters) and instrument-specific transformations. Our method uses the common structure of many observations, in which the same underlying physical object (e.g. a star) is observed repeatedly under different sensor configurations, and the same sensor observes multiple physical systems. This data structure enables a novel form of contrastive supervision.

\paragraph{Triplet-based contrastive learning.} We leverage the causal structure of astronomical observations through triplet-based contrastive learning. Each training example consists of three observations:
$(F^{(\text{anchor})}, F^{(\text{same\_star})}, F^{(\text{same\_inst})}),$
where the anchor and same-star observations share stellar identity but differ in instrumentation, while the anchor and same-instrument observations share instrumentation but differ in stellar identity. This triplet construction directly encodes our modeling assumptions: stellar latent representations should be invariant across instrument changes, while instrumental representations should be invariant across stars.

\paragraph{Architecture.}The architecture of our model is shown in Fig.~\ref{fig:model-architecture}. Given an observed time series $F_{\text{observed}}^{(n)}(t)$ generated from star $s$ observed by instrument $m$, the model learns two latent representations: a stellar encoding $\bm{z}_{\text{star}}^{(n)}$ and an instrumental encoding $\bm{z}_{\text{instr}}^{(n)}$. Both are produced from the same input via two separate encoders with identical architecture but untied parameters:
\begin{equation}
\bm{z}_{\text{star}}^{(n)} = E_{\text{star}}(F^{(n)}, t), \quad \bm{z}_{\text{instr}}^{(n)} = E_{\text{instr}}(F^{(n)}, t).
\end{equation}
Each encoder follows a Conformer-based architecture \citep{gulati2020}, incorporating temporal self-attention, LSTM recurrence, and depthwise convolution to capture patterns across a range of timescales. The outputs are globally pooled and projected into a fixed-dimensional latent space (see Appendix \ref{app:transformer} for details). To reconstruct the original input, the decoder first projects each latent representation into a shared hidden space via learned nonlinear transformations, and then combines them multiplicatively:
\begin{equation}
\hat{F}_{\text{observed}}^{(n)} = D\left( \bm{g}_{\text{star}} \odot \bm{g}_{\text{instr}} \right),
\end{equation}
where  $\bm{g}_{\text{star}}$ and $\bm{g}_{\text{instr}}$ are projections of the respective latent parameters $\bm{z}_{\text{star}}$ and $\bm{z}_{\text{instr}}$ mapped a small MLP projection head, and $\odot$ denotes element-wise multiplication (Hadamard product). The decoder then maps the fused embedding back to the time series domain.

\paragraph{Contrastive objectives.} To encourage disentanglement, we apply contrastive losses in both latent spaces. For each anchor embedding $\mathbf{a}$, we define sets of positives $\mathcal{P}(a)$ and negatives $\mathcal{N}(a)$ based on metadata: for the stellar latent space, positives are light curves of the same star observed under different instruments; for the instrument latent space, positives are from the same instrument observing different stars.

We minimize a generalized InfoNCE loss $\mathcal{L}_{\text{InfoNCE}}(\mathbf{a})$:
\begin{equation}
%\mathcal{L}_{\text{InfoNCE}}(\mathbf{a}) = 
-\log \frac{ \sum_{\mathbf{p} \in \mathcal{P}(a)} \exp(\mathbf{a}^\top \mathbf{p} / \tau)}{ \sum_{\mathbf{p} \in \mathcal{P}(a)} \exp(\mathbf{a}^\top \mathbf{p} / \tau) + \sum_{\mathbf{n} \in \mathcal{N}(a)} \exp(\mathbf{a}^\top \mathbf{n} / \tau)},
\end{equation}
where $\tau$ is a temperature parameter and all vectors are $\ell_2$ normalized. We compute this loss across all anchor samples in the batch for both latent spaces:
\begin{align}
\mathcal{L}_{\text{star}} &= \frac{1}{B} \sum_{n=1}^{B} \mathcal{L}_{\text{InfoNCE}}(\bm{p}_{\text{star}}^{(n)}), \\
\mathcal{L}_{\text{instr}} &= \frac{1}{B} \sum_{n=1}^{B} \mathcal{L}_{\text{InfoNCE}}(\bm{p}_{\text{instr}}^{(n)}),
\end{align}
where $B$ is the number of anchor samples in the training batch, and $\bm{p}_{\text{star}}$ and $\bm{p}_{\text{instr}}$ are projections of the respective latent parameters $\bm{z}_{\text{star}}$ and $\bm{z}_{\text{instr}}$ mapped by a small MLP projection head as described in \citet{Chen2020}.

This formulation generalizes the InfoNCE loss used in SimCLR \citep{Chen2020} to allow multiple positives per anchor selected via structured metadata.

\paragraph{Full objective.} We train the model with a weighted sum of reconstruction and contrastive objectives:
\begin{equation}
\mathcal{L}_{\text{total}} = \lambda_{\text{recon}} \mathcal{L}_{\text{recon}} + \lambda_{\text{star}} \mathcal{L}_{\text{star}} + \lambda_{\text{instr}} \mathcal{L}_{\text{instr}},
\end{equation}
where the reconstruction loss is defined as a masked mean squared error:
\begin{equation}
\mathcal{L}_{\text{recon}} = \frac{1}{T} \sum_{t=1}^{T} \left( \hat{F}_{\text{observed}}^{(n)}[t] - F_{\text{observed}}^{(n)}[t] \right)^2.
\end{equation}
At inference time, the projection heads used for contrastive training are discarded. Downstream tasks operate on the disentangled latent representations, which we demonstrate are interpretable and predictive of physical parameters, particularly in low-data regimes.

\section{Results \& Discussion}
\label{sect:results}

We evaluate our causal foundation model by exploring the quality of its learned latent spaces and its performance on downstream predictive tasks. We compare against a foundation model baseline with identical architecture but a single shared latent space, trained with contrastive learning over same-star pairs only (no instrumental disentanglement).

\subsection{Quality of the embedding}
\label{subsect:quality}

To assess whether the model successfully disentangles stellar and instrumental factors, we analyze the learned latent spaces $\bm{z}_{\text{star}}$ and $\bm{z}_{\text{instr}}$ using UMAP \citep{McInnes2018} and their correlation with ground-truth parameters. Each embedding is projected to 2D, and color-coded by either instrument ID ($m$) or stellar properties ($\theta_{s,0}$).

\begin{figure}[ht]
\vskip 0.2in
%\vskip -0.1in
\begin{center}
\centerline{\includegraphics[width=\columnwidth]{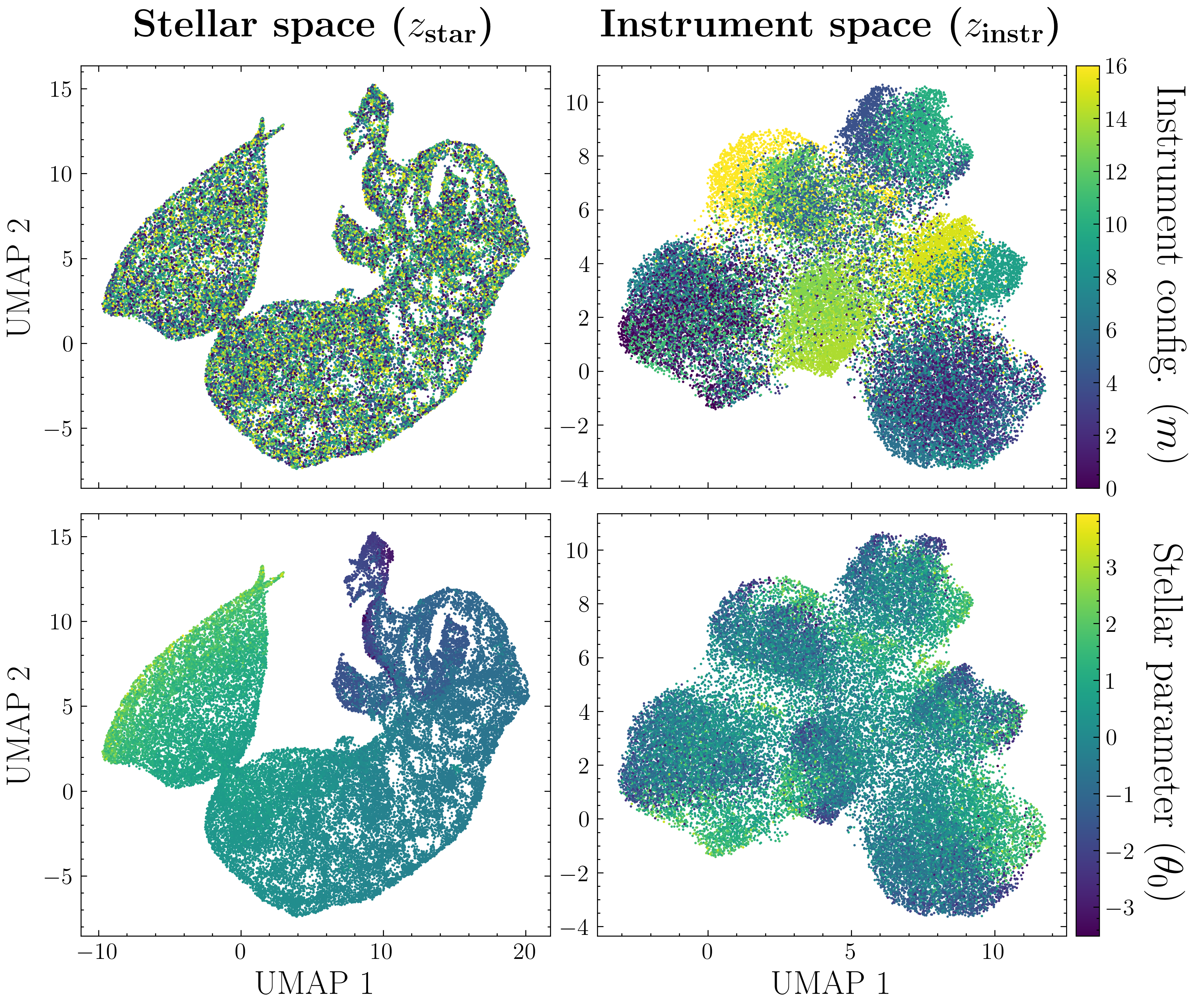}}
%\vskip -0.1in
\caption{UMAP projections of stellar ($\bm{z}_{\text{star}}$, left) and instrumental ($\bm{z}_{\text{instr}}$, right) latent spaces. Top: colored by instrument configuration. Bottom: colored by primary stellar parameter $\theta_{s,0}$. The stellar space captures physical properties while minimizing instrument clustering. The instrument space reveals strong sector structure but shows partial leakage of stellar information.
}
\label{fig:umap_sector}
\end{center}
\vskip -0.2in
\end{figure}

As illustrated in Fig.~\ref{fig:umap_sector}, the stellar latent space exhibits strong alignment with intrinsic physical property $\theta_{s,0}$ and shows minimal clustering by instrument. In contrast, the instrument latent space is well-structured by instrument configuration $m$, as expected. However, we also observe a moderate correlation between $\theta_{s,0}$ and the instrument latent space, indicating some leakage of physical information into $\bm{z}_{\text{instr}}$.

It may be that the recorded brightness can exhibit dependencies that couple stellar and instrumental properties (e.g., brighter stars may be more affected by certain instrumental distortions). In future work, we will explore minimizing the mutual information between the latent spaces to improve the seperations.

\subsection{Downstream tasks}
\label{subsect:tasks}

To assess the utility of the learned representations, we evaluate performance on a supervised downstream task: predicting the primary stellar parameter $\theta_{s,0}$ from limited labeled data. We train a lightweight MLP regressor using four different input representations: (i) raw light curves (baseline), (ii) latent embeddings from a baseline foundation model with a single shared latent space trained using a contrastive loss on same-star pairs, and (iii) our proposed disentangled representations from the stellar latent space ($\bm{z}_{\text{star}}$) and (iv) the instrument latent space ($\bm{z}_{\text{instr}}$).

\begin{figure}[ht]
\vskip 0.2in
\begin{center}
\centerline{\includegraphics[width=\columnwidth]{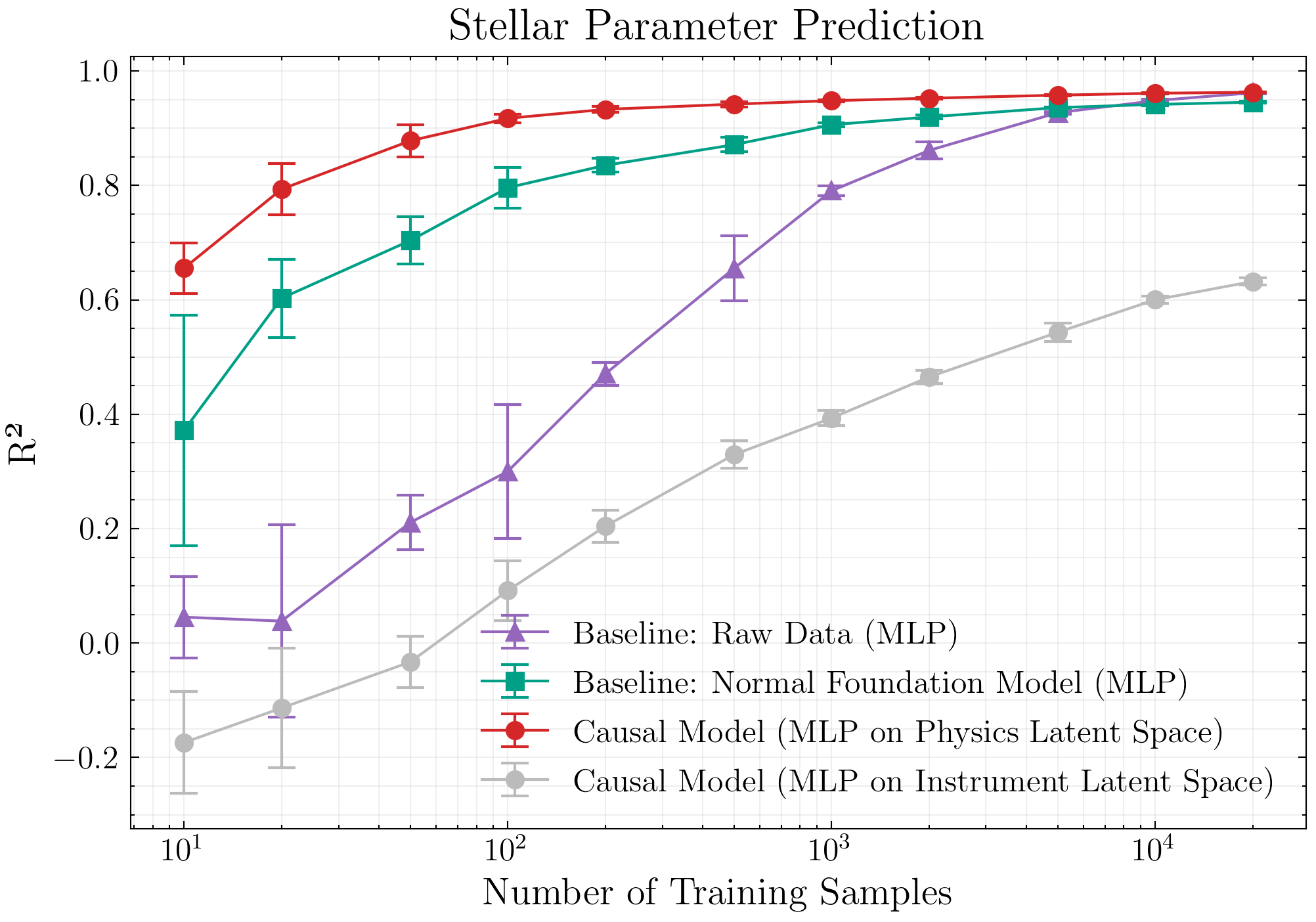}}
%\vskip -0.1in
\caption{Comparison of prediction performance across limited training sample sizes. The plots show the $R^2$ for predicting the stellar parameter using four approaches: MLP on raw light curves (baseline), MLP on a baseline foundation model (single latent space with contrastive loss on same-star pairs), MLP on the stellar latent space, and MLP on the instrument latent space. Data points indicate mean and standard deviation across five evaluation runs. The latent space-based methods demonstrate superior performance, especially with limited training data.
}
\label{fig:downstream-tasks}
\end{center}
\vskip -0.2in
\end{figure}

Fig.~\ref{fig:downstream-tasks} shows the $R^2$ scores between the predicted and ground-truth stellar parameters for each input across different training set sizes. Models trained on $\bm{z}_{\text{star}}$ consistently outperform those using raw data or baseline latent spaces, especially in the few-shot regime. The model trained on our stellar or physics latent space performs as well or better than the normal foundation model with ten times less training data. These results demonstrate that our model captures meaningful stellar properties in its latent space, enabling effective few-shot learning and strong generalization from unlabeled, heterogeneous observational data to predictive downstream tasks.

\section{Conclusions}

Our results suggest that incorporating causal structure into foundation models--specifically through a dual-latent design that separately encodes physical and instrumental components--can significantly outperform conventional foundation models with a single latent space, particularly in limited-data regimes. The model’s use of structured triplet-based contrastive learning enables effective disentanglement of generative factors, leveraging observational metadata rather than requiring explicit supervision. Although our results are based on a conformer encoder, we observed similar results with simpler MLP-based architectures, suggesting that the disentanglement is due to the contrastive learning objective rather than architectural complexity.

Beyond astronomy, this methodology is broadly applicable to domains where observational confounding is a concern and where structured triplet relationships can be inferred or constructed, such as biomedicine, remote sensing, or climate forecasting. Our preliminary experiments on NASA TESS light curves demonstrate the method’s potential for real-world application. Future work will explore improved disentanglement strategies, deployment on large mission-scale datasets, and extensions to multimodal and cross-domain foundation models.

%\section*{Disclaimer} As requested by the submission guidelines, we note that this work has also been accepted to the Foundation Models for Structured Data (FMSD) Workshop at ICML 2025. 

%\clearpage

\section*{Software and Data}

We used Pytorch and JAX to develop the model and simulations and will make the full code publicly available after acceptance.

%Acknowledgements should only appear in the accepted version.
\section*{Acknowledgements}
Funding for the TESS and Kepler missions is provided by NASA's Science Mission Directorate. The Villar Astro Time Lab acknowledges support through the David and Lucile Packard Foundation, National Science Foundation under AST-2433718, AST-2407922 and AST-2406110, as well as an Aramont Fellowship for Emerging Science Research. This work is supported by the National Science Foundation under Cooperative Agreement PHY-2019786 (The NSF AI Institute for Artificial Intelligence and Fundamental Interactions, http://iaifi.org/). 

\section*{Impact Statement}

This paper presents work whose goal is to advance the field of machine learning, signal processing and astrophysics. There are many potential societal consequences of our work, none which we feel must be specifically highlighted here.

\bibliography{example_paper}
\bibliographystyle{icml2025}

%%%%%%%%%%%%%%%%%%%%%%%%%%%%%%%%%%%%%%%%%%%%%%%%%%%%%%%%%%%%%%%%%%%%%%%%%%%%%%%
%%%%%%%%%%%%%%%%%%%%%%%%%%%%%%%%%%%%%%%%%%%%%%%%%%%%%%%%%%%%%%%%%%%%%%%%%%%%%%%
% APPENDIX
%%%%%%%%%%%%%%%%%%%%%%%%%%%%%%%%%%%%%%%%%%%%%%%%%%%%%%%%%%%%%%%%%%%%%%%%%%%%%%%
%%%%%%%%%%%%%%%%%%%%%%%%%%%%%%%%%%%%%%%%%%%%%%%%%%%%%%%%%%%%%%%%%%%%%%%%%%%%%%%
\newpage
\appendix
\onecolumn
\section{Simulated data}

\begin{figure}[h]
\vskip 0.2in
    \centering
    \includegraphics[width=0.8\linewidth]{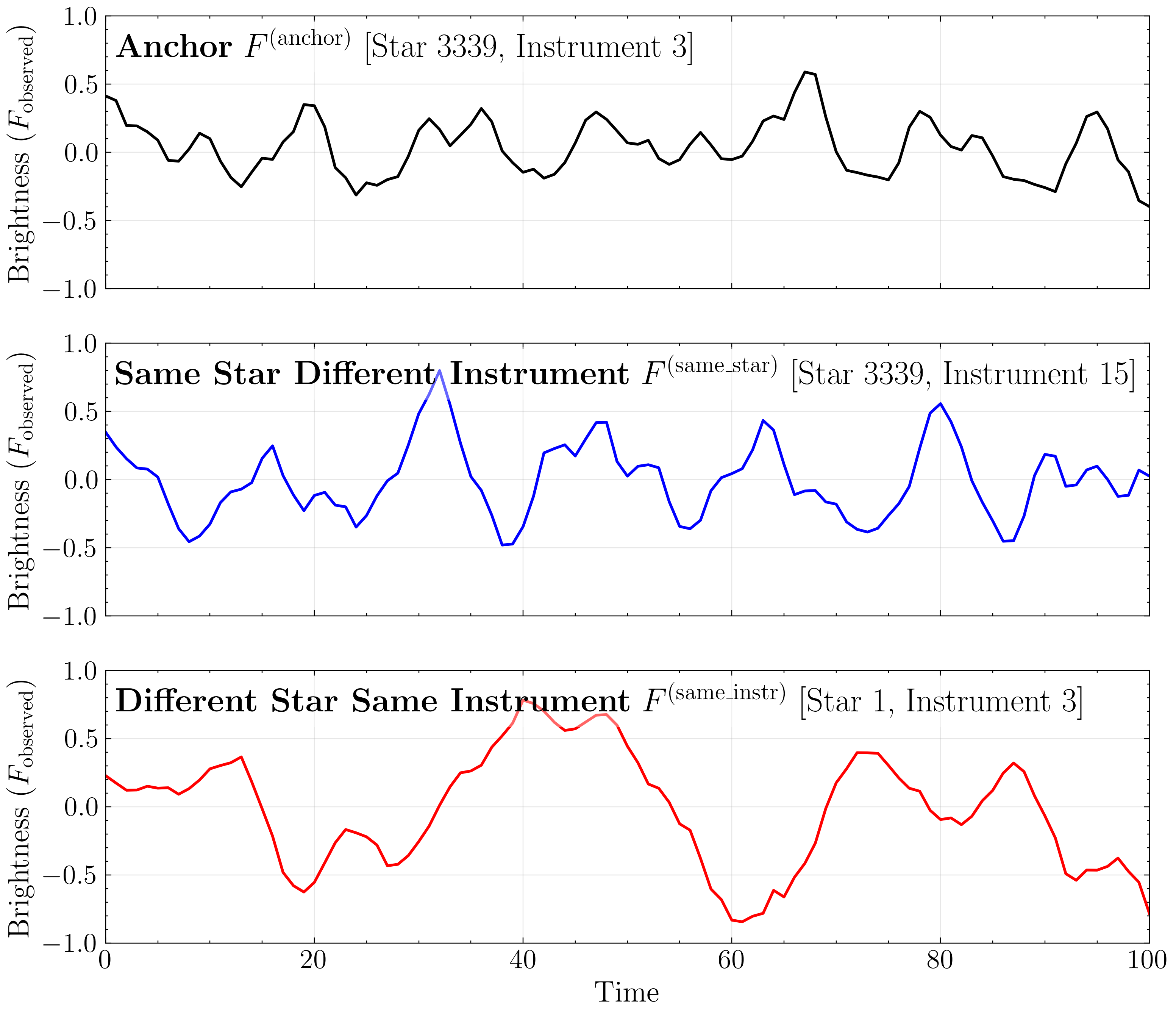}
    \caption{Example of simulated triplet data, as explained in Section.~\ref{sect:data}.}
    \label{fig:example-lc}
\end{figure}

We show an example triplet of our simulated time series in Fig.~\ref{fig:example-lc}. The top row is the anchor, middle row the same star but observed by different instrument and the bottom row a random star observed by the same instrument as the anchor.

\clearpage
\section{Encoder architecture}
\label{app:transformer}
\begin{figure}[ht]
\vskip 0.2in
\begin{center}
\centerline{\includegraphics[width=0.3\columnwidth]{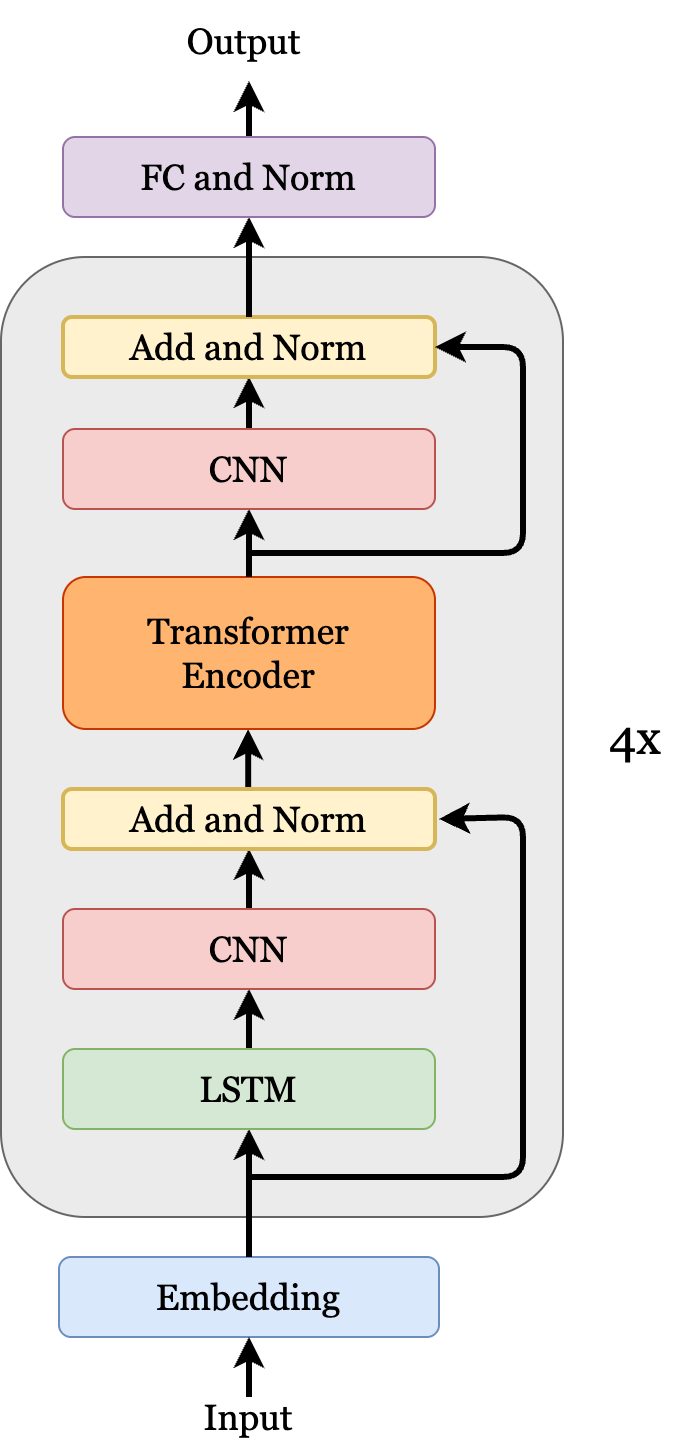}}
\caption{Encoder architecture.}
\label{fig:transformer-architecture}
\end{center}
\vskip -0.2in
\end{figure}

The detailed encoder architecture is shown in Fig.~\ref{fig:transformer-architecture}. The time series encoding architecture was originally developed as a time series classifier for the TESS mission (Gregory et al., in prep.), inspired by the conformer architecture \cite{gulati2020} for astronomical data from \citet{Pan2024}. 

\textbf{Implementation Details:} Each encoder (as illustrated in Fig.~\ref{fig:transformer-architecture}) consists of 4 conformer blocks with 64-dimensional embeddings, 4 attention heads, and 128-dimensional feed-forward networks. The stellar and instrumental latent spaces each have dimensionality 20. We train with Adam optimizer (learning rate=0.001), early-stopping based on the validation loss, and loss weights $\lambda_{\text{recon}}=1.0$, $\lambda_{\text{star}}=1.0$, $\lambda_{\text{instr}}=1.0$.

%%%%%%%%%%%%%%%%%%%%%%%%%%%%%%%%%%%%%%%%%%%%%%%%%%%%%%%%%%%%%%%%%%%%%%%%%%%%%%%
%%%%%%%%%%%%%%%%%%%%%%%%%%%%%%%%%%%%%%%%%%%%%%%%%%%%%%%%%%%%%%%%%%%%%%%%%%%%%%%

\end{document}